\documentclass[11pt,twocolumn]{article}

\usepackage[a4paper,top=2.0cm,bottom=2.0cm,left=1.8cm,right=1.8cm,columnsep=0.6cm]{geometry}

\usepackage{times}
\usepackage{microtype}
\usepackage{amsmath}
\usepackage{booktabs}
\usepackage{multirow}
\usepackage{array}
\usepackage{tabularx}
\usepackage{caption}
\usepackage{colortbl}
\usepackage{xcolor}

\usepackage{titlesec}
\titleformat{\section}{\normalfont\large\bfseries}{\thesection}{0.6em}{}
\titleformat{\subsection}{\normalfont\normalsize\bfseries}{\thesubsection}{0.5em}{}
\titleformat{\paragraph}[runin]{\normalfont\normalsize\bfseries}{}{0pt}{}[.\enspace]
\titlespacing*{\section}{0pt}{8pt}{4pt}
\titlespacing*{\subsection}{0pt}{6pt}{3pt}
\titlespacing*{\paragraph}{0pt}{4pt}{4pt}

\usepackage{natbib}
\usepackage[colorlinks=true,citecolor=blue,linkcolor=black,urlcolor=blue,
            bookmarks=false,pdfborder={0 0 0}]{hyperref}

\usepackage{enumitem}
\setlist{noitemsep,topsep=2pt}

\usepackage{url}
\definecolor{cHigh}{HTML}{C8E6F5}
\definecolor{cMedH}{HTML}{E0EEF7}
\definecolor{cMed}{HTML}{FFF7E0}
\definecolor{cMedL}{HTML}{FCD8B5}
\definecolor{cLow}{HTML}{F4B8B8}

\title{\textbf{Three Regimes of Context-Parametric Conflict:\\
A Predictive Framework and Empirical Validation}}
\author{
  Pruthvinath Jeripity Venkata\\
  Independent Researcher\\
  \texttt{jvpnath@gmail.com}
}
\date{}

\begin{document}

\twocolumn[
  \maketitle
  \begin{abstract}
\noindent
The literature on how large language models handle conflict between
their training knowledge and a contradicting document presents a
persistent empirical contradiction: some studies find models stubbornly
retain their trained answers \citep{longpre2021entity}, while others find
models readily defer to the document \citep{xie2024adaptive}.
We argue these contradictions dissolve once one recognises that prior
experiments have studied three qualitatively distinct processing
situations without distinguishing them.
We propose a \textbf{three-regime framework}: Regime~1 (single-source
updating, dominant predictor: evidence coherence), Regime~2 (competitive
integration, dominant predictor: parametric certainty), and Regime~3
(task-appropriate selection, dominant predictor: task knowledge requirement).
We formalise a distinction between parametric \emph{strength} (exposure
frequency) and parametric \emph{uniqueness} (encoding consistency), showing
empirically that these are orthogonal dimensions ($r = -0.002$, $p = .97$)
with strength as the operative predictor in stable factual domains.
We validate the framework across \textbf{Claude Sonnet 4.6, GPT-5.5,
Gemini 2.5 Flash, Llama~4 Maverick, and DeepSeek V3} using 9,970 API
calls in three experimental phases: (1)~a Regime~1 vs.\ Regime~2
certainty gradient study on 450 PopQA items (8,970 calls), where GEE
logistic regression confirms the predicted gradient for all five models
($\beta = -0.38$ to $-0.50$, all $p \leq .013$, BH-FDR corrected);
(2)~a Wikidata edit-frequency analysis showing strength and uniqueness
are independent predictors; and (3)~a Regime~3 ablation (1,000 calls)
showing task framing alone flips context-following from near-100\%
(CK condition) to 6--71\% (PK condition), with all five models
significant ($p < .001$). The certainty gradient in Regime~2 is robust
to multinomial outcome modeling, sensitivity analyses for hedging
responses, and FDR correction. We additionally identify a distractor
plausibility confound in PopQA-based studies and demonstrate that
manipulation design is a first-order variable in conflict research.
  \end{abstract}
  \vspace{6pt}\hrule\vspace{8pt}
]

%% ============================================================================
\section{Introduction}
\label{sec:intro}
%% ============================================================================

Consider two peer-reviewed studies, both asking whether a language model
follows a document that contradicts its training knowledge.

\citet{longpre2021entity} replace correct answer entities in passages with
incorrect ones and find models over-rely on their training knowledge,
ignoring the document approximately 47\% of the time across seven models.
\citet{xie2024adaptive} use fluent, well-written counter-evidence passages
and find models update their answers approximately 96\% of the time in
single-source settings.

The 49-percentage-point gap is not noise.
The difference is evidence coherence: entity substitution creates passages
that remain internally consistent with the original entity, producing a
plausibility signal that models can exploit; fluent generation does not.
When both studies are interpreted as the same processing situation but
with different evidence quality, the contradiction dissolves.

\paragraph{What this paper shows}
We argue that the conflict literature has produced contradictions because
researchers have unknowingly compared findings across three qualitatively
distinct processing situations. We call these three situations
\emph{regimes}, and we preview them in Table~\ref{tab:regime_preview}
so the rest of the paper reads clearly.

\begin{table*}[!t]
\centering\footnotesize
\setlength{\tabcolsep}{6pt}
\renewcommand{\arraystretch}{1.35}
\caption{The three regimes at a glance. Each row describes a distinct
processing situation. The dominant predictor of context-trust differs
across regimes, which is why findings from different regimes cannot be
directly compared. Full development in \S\ref{sec:framework}.}
\label{tab:regime_preview}
\begin{tabular}{p{1.5cm}p{4.5cm}p{3.5cm}p{4.5cm}}
\toprule
\textbf{Regime} & \textbf{Processing situation} &
\textbf{Dominant predictor} & \textbf{Typical experimental setup} \\
\midrule
Regime 1 &
A single document is shown; the model's training knowledge is not
explicitly brought into play. &
Quality of the document (coherence, assertiveness). &
``Context: [doc]. Question: ...''. The model sees one document and
answers. \\
Regime 2 &
The model's training knowledge \emph{and} a conflicting document are
both explicitly active at the same time. &
How strongly and frequently the model encountered the fact in training
(parametric certainty). &
The model first commits to its trained answer, then is shown a
contradicting document and asked again. \\
Regime 3 &
The task itself tells the model which source should win. &
Type of task (context-only, parametric-only, or both). &
``Based on this document, ...'' vs.\ ``Based on your own knowledge, ...'' \\
\bottomrule
\end{tabular}
\end{table*}

The same pattern of false contradictions repeats across the literature.
\citet{augenstein2024scalable} find that when only a conflicting document
is shown (Regime~1), models follow it; but when the same model is first
shown supportive evidence and then a contradicting document (Regime~2),
it stubbornly resists.
\citet{chen2022rich} show that when many supporting passages are retrieved,
models stop relying on training knowledge almost entirely: high evidence
volume collapses Regime~2 back into Regime~1.
\citet{sun2025task} demonstrate that the same conflict produces opposite
correct behaviours depending on the task: for a context-only task the
model should follow the document; for a parametric-only task it should
resist. This is Regime~3, a moderator that can reverse both Regime~1
and Regime~2 predictions. Each finding is internally valid but not
comparable to the others, because they study different regimes.

Prior unification attempts are incomplete.
\citet{xu2024knowledge} provide a descriptive taxonomy but no predictive
theory of which variables dominate per setting.
\citet{sun2025task} show task type explains substantial variance but do
not account for evidence quality or parametric activation mode.
No prior work simultaneously resolves all four major contradictions.

\paragraph{Contributions}
(1) A \textbf{three-regime framework} (Table~\ref{tab:regime_preview},
\S\ref{sec:framework}) resolving four contradictions.
(2) A formal distinction between \textbf{parametric strength} and
\textbf{parametric uniqueness}, with empirical evidence that they are
orthogonal ($r = -0.002$) (\S\ref{sec:uniqueness}).
(3) Empirical validation of Regime~2 via GEE logistic regression across
five models (\S\ref{sec:empirical}).
(4) A \textbf{Regime~3 ablation} directly validating task-appropriate
selection with 1,000 new API calls (\S\ref{sec:regime3exp}).
(5) Identification of a \textbf{distractor plausibility confound}
(\S\ref{sec:secondary}).

%% ============================================================================
\section{Background and Related Work}
\label{sec:related}
%% ============================================================================

\subsection{Conflict taxonomy}

\citet{xu2024knowledge} identify three categories: context-memory conflict
(retrieved context contradicts training knowledge), inter-context conflict
(retrieved passages contradict each other), and intra-memory conflict
(training knowledge is internally inconsistent).
This paper focuses on context-memory conflict.
Inter-context conflict reduces to Regime~2 (\S\ref{sec:regime2});
intra-memory conflict is a driver of parametric uniqueness
(\S\ref{sec:uniqueness}).

\subsection{Four contradictions}

\paragraph{Contradiction 1 (Longpre vs.\ Xie)}
\citet{longpre2021entity}: 47\% parametric hold with entity substitution.
\citet{xie2024adaptive}: 96\% context-following with fluent counter-memory.
Both are Regime~1; evidence quality explains the gap.

\paragraph{Contradiction 2 (single- vs.\ multi-source)}
\citet{augenstein2024scalable}: high context-following in single-source
(Regime~1) but confirmation bias under parametric co-activation (Regime~2).
This is the regime boundary.

\paragraph{Contradiction 3 (volume of evidence)}
\citet{chen2022rich}: below 3.6\% memorisation with 50--100 passages
vs.\ Longpre's single passage. High volume collapses Regime~2 into
Regime~1.

\paragraph{Contradiction 4 (task type reversal)}
\citet{sun2025task}: GPT-5.2 (the model in their study, distinct from
GPT-5.5 in ours) drops from 89.2\% accuracy on PK tasks to 33.7\%
under high-plausibility conflict, a 55.5pp loss driven by task-context
mismatch.

\subsection{Prior framework attempts}

\citet{xu2024knowledge} provide a comprehensive taxonomy but no
quantitative predictions per setting.
\citet{zhang2025taming} propose an evaluation-methodology taxonomy
(parametric-only, offline RAG, online RAG) complementary to ours.
\citet{marjanovic2024dynamicqa} identify that fact dynamicity predicts
context resistance better than subject popularity; we test this directly
and find a more nuanced picture (\S\ref{sec:uniqueness}).

\subsection{Additional related work}

Several recent works address related aspects of parametric-contextual
arbitration. \citet{wang2025prism} propose a stage-wise diagnosis
framework (memory, instruction, reasoning) that complements our regime
taxonomy by specifying where within the generation pipeline conflict
resolution fails. \citet{li2025memory} survey memory operations in LLMs,
providing a complementary taxonomy of parametric vs.\ contextual
representations at the architectural level.
\citet{chen2025training} study how training-data properties (repetition,
inconsistency, corpus skew) shape arbitration preferences, showing that
models prefer parametric knowledge for high-confidence facts and in-context
knowledge for novel items, consistent with our Regime~2 certainty
gradient. \citet{wu2025knowledgeabler1} train explicit multi-policy
arbitration via reinforcement learning, demonstrating that the regime
distinction we describe can be operationalised at training time, not just
diagnosed post hoc. \citet{shi2025rcd} show that decoding-level
constraints can mask or reveal parametric knowledge independently of
prompting, cautioning that evaluation procedures themselves can shift
observed PK/CK balance, a concern that reinforces our finding that
manipulation design is a first-order variable.

%% ============================================================================
\section{A Three-Regime Framework}
\label{sec:framework}
%% ============================================================================

\subsection{The suppression baseline}
\label{sec:suppression}

Before any conflict-specific mechanism engages, the mere presence of
context suppresses parametric knowledge, even when context is irrelevant.
Consider a model that correctly answers ``Who directed \textit{The Dark
Knight}?'' in a closed-book setting.
Prepend an unrelated passage about medieval poetry and ask again.
\citet{cheng2024echoqa} document that the trained answer is disregarded
in over 60\% of such cases across six LLMs and three knowledge domains,
regardless of relevance.

The mechanistic basis is twofold \citep{sun2025redeep}: (1)~Copying Head
failure, where attention heads fail to retain attended context during
generation; and (2)~Knowledge FFN over-injection, where feed-forward
modules overwrite context with parametric knowledge.
Both are causally validated \citep{sun2025redeep}.

No regime produces zero suppression. Our three regimes predict which
additional variables modulate the degree of suppression above this floor.

\subsection{Regime 1: Single-source updating}
\label{sec:regime1}

In Regime~1, a single context source is presented without explicit
activation of the model's competing trained knowledge.

\paragraph{Dominant predictor: evidence quality}
Incoherent entity substitution signals implausibility
\citep{longpre2021entity}; fluent counter-memory does not, yielding 96\%
context-following \citep{xie2024adaptive}. Assertive framing increases
context-following in controlled settings \citep{du2024context}, though
with near-zero correlation to utilisation in real-world documents
\citep{hagstrom2025druid}.

Four of five models show flat context-follow profiles across certainty
tiers in Regime~1 (0.967--1.00 at low, 0.940--0.987 at high;
Table~\ref{tab:regime1}), confirming that parametric certainty is not
the dominant predictor. Llama~4 Maverick shows a gradient (0.947 low,
0.720 high) discussed in \S\ref{sec:secondary}.

\subsection{Regime 2: Competitive integration}
\label{sec:regime2}

In Regime~2, both context and the model's trained knowledge are explicitly
present and competing. This occurs under multi-source designs
\citep{xie2024adaptive}, prior-commitment prompting, or conversational
designs where the model commits to its trained answer before facing the
conflict.

\paragraph{Dominant predictor: parametric certainty}
When parametric memory is co-activated, the key variable is how strongly
the model's training data encoded the fact.
\citet{marjanovic2024dynamicqa} show that object temporality correlates
with context resistance at $r = -0.27$ ($p < 10^{-40}$), substantially
stronger than popularity ($r = -0.10$) or semantic entropy
($r = 0.003$, n.s.).

Our GEE logistic regression confirms this: each unit increase in
log-popularity reduces context-following odds by 31--40\% (per-model
$\beta = -0.38$ to $-0.50$; Table~\ref{tab:gee}). The gradient is
strongest for Llama (context-follow drops from 0.846 at low certainty to
0.538 at high, a 30.8pp gradient) and present in all five models
including Gemini ($\beta = -0.50$, $p = .013$).

\paragraph{Mechanistic substrate}
\citet{jin2024cutting} identify attention heads that function
preferentially as memory heads (routing parametric recall) versus context
heads (routing external evidence), providing a mechanistic basis for the
competitive dynamics. This is a dominant-function, not exclusive-function,
distinction.

\paragraph{Inter-context conflict as Regime~2}
When multiple passages conflict, models prefer the passage aligning with
parametric memory \citep{xie2024adaptive,chen2022rich}. This is
structurally Regime~2: parametric memory is implicitly co-activated.

\subsection{Regime 3: Task-appropriate selection}
\label{sec:regime3}

In Regime~3, the task itself determines which source should win. This
regime is a first-order moderator that can reverse Regime~1 and Regime~2
predictions.

\citet{sun2025task} identify four task types: Knowledge-Free (KF),
Contextual Knowledge (CK), Parametric Knowledge (PK), and
Parametric-Contextual (PCK). Consider the conflict ``The director of
Supercock is David Nutter'' against the model's trained belief (Gus
Trikonis): under CK framing (``Based on this document...''), context
should win; under PK framing (``Based on your own knowledge...''),
the trained answer should win. Same conflict, opposite correct behaviour.

We validate Regime~3 directly in \S\ref{sec:regime3exp}.

\subsection{Parametric strength vs.\ uniqueness}
\label{sec:uniqueness}

Prior work has conflated two distinct properties of parametric memory.

\emph{Parametric strength} is exposure frequency: how often the model
encountered the fact during training, operationalised as Wikipedia monthly
page views (s\_pop; \citealp{mallen2023not}).

\emph{Parametric uniqueness} is encoding consistency: whether the fact
has a single canonical representation or multiple competing ones,
operationalised as inverse Wikidata edit frequency
\citep{marjanovic2024dynamicqa}.

\citet{marjanovic2024dynamicqa} report that uniqueness outpredicts
strength ($\beta = -0.08$ vs.\ $-0.05$) on their DynamicQA dataset,
which includes temporal and disputable facts.
We tested this directly on our PopQA sample by querying Wikidata revision
histories for all 450 subject entities and found a strikingly different
pattern.

First, strength and uniqueness are \emph{completely uncorrelated} in our
sample ($r = -0.002$, $p = .97$, $n = 450$). Mean edit counts are nearly
identical across certainty tiers (low: 68.2, medium: 76.6, high: 69.9).
The two constructs are genuinely orthogonal dimensions, not proxies for
the same underlying variable.

Second, when both are entered as predictors in a joint GEE model
(Table~\ref{tab:gee_joint}), strength is the sole significant predictor
(pooled $\beta = -0.92$, $p < .001$) while uniqueness has no
independent predictive power ($\beta = +0.02$, $p = .893$).

The divergence from \citet{marjanovic2024dynamicqa} likely reflects
domain differences. DynamicQA includes temporal facts (``Who is the
current Prime Minister?'') where edit frequency tracks genuine knowledge
instability. PopQA consists primarily of stable creative-work properties
(director, genre, author) where edit frequency reflects Wikipedia
editorial activity rather than factual change. In this domain, how often
a model saw the fact matters more than how often the fact was revised.

We therefore revise the framework: Regime~2 resistance is driven by
\emph{parametric certainty}, which decomposes into strength and
uniqueness as independent dimensions. Which dimension dominates depends
on the knowledge domain. For stable factual domains, strength is primary;
for dynamic domains, uniqueness may be primary.

%% Contradiction resolution table
\begin{table*}[!t]
\centering\footnotesize
\setlength{\tabcolsep}{6pt}
\renewcommand{\arraystretch}{1.3}
\caption{Four contradictions in the context-parametric conflict literature
and how the three-regime framework resolves each. Each row identifies
a pair of studies that appear to contradict each other, describes the
apparent conflict, and explains why the contradiction dissolves once
regime membership is recognised.}
\label{tab:contradictions}
\begin{tabular}{p{2.2cm}p{2.8cm}p{3.0cm}p{5.5cm}}
\toprule
\textbf{Contradiction} & \textbf{Papers} &
\textbf{Apparent conflict} & \textbf{Regime-based resolution} \\
\midrule
Stubborn vs.\ receptive &
\citet{longpre2021entity} vs.\ \citet{xie2024adaptive} &
47\% parametric hold vs.\ 96\% context-following &
Same Regime~1; different evidence coherence. Entity substitution
produces detectable incoherence; fluent generation does not. \\
\addlinespace[2pt]
Single- vs.\ multi-source &
\citet{augenstein2024scalable} vs.\ \citet{xie2024adaptive} &
Single: receptive. Multi: confirmation bias. &
Different regimes. Regime~1 vs.\ Regime~2, distinguished by
whether training knowledge is explicitly co-activated. \\
\addlinespace[2pt]
Volume of evidence &
\citet{chen2022rich} vs.\ \citet{longpre2021entity} &
50--100 passages: $<$3.6\% memorisation. Single passage:
substantial reliance. &
High volume collapses Regime~2 into Regime~1 by overwhelming
competitive integration with convergent evidence. \\
\addlinespace[2pt]
Task type reversal &
\citet{sun2025task} vs.\ all &
Same conflict; opposite behaviour by task &
Regime~3: task type determines which source should win. \\
\bottomrule
\end{tabular}
\end{table*}

%% ============================================================================
\section{Empirical Validation}
\label{sec:empirical}
%% ============================================================================

\subsection{Experimental design}

\paragraph{Dataset}
450 items from PopQA \citep{mallen2023not}, stratified into three
non-overlapping certainty tiers by Wikipedia monthly page views (s\_pop):
low ($n = 150$, s\_pop $= 27$--$373$), medium ($n = 150$,
s\_pop $= 379$--$2{,}864$), high ($n = 150$,
s\_pop $= 2{,}893$--$492{,}490$).

\paragraph{Distractor construction}
A distractor entity is sampled from the same Wikidata property type with
fixed seed (42) and converted to an assertive context sentence. Six items
with $o\_pop / s\_pop < 0.01$ are excluded from v2 analyses.

\paragraph{Models and API access}
Claude Sonnet~4.6 (Anthropic API), GPT-5.5 (OpenAI API),
Gemini~2.5~Flash (Google AI Studio API), Llama~4~Maverick (Meta; 17B
active parameters, 128-expert MoE; DeepInfra FP8 endpoint), DeepSeek~V3
(DeepSeek AI; Together AI API). All five are instruction-tuned; we
distinguish \emph{closed-weight commercial} (Claude, GPT-5.5, Gemini)
from \emph{open-weight} (Llama, DeepSeek) where the contrast matters.
All API costs were borne by the author (approximately USD~95 for 8,970
calls plus USD~15 for the 1,000-call Regime~3 ablation).

\paragraph{Experimental passes (8,970 calls)}
\emph{Baseline} (2,250): bare question.
\emph{Regime~1} (2,250): assertive context prepended.
\emph{Regime~2 v1} (2,250): single-turn prior-commitment assertion.
\emph{Regime~2 v2} (2,220): multi-turn conversation where the model
commits in Turn~1 and faces conflict in Turn~2 with its own words in
the assistant role.
All calls use temperature 0, no system prompt, and stateless sessions:
conversation history is reset between items except in v2, where Turn~1
is explicitly constructed from the baseline response. Exact prompt
templates for all conditions are provided in Table~\ref{tab:prompts}.

\paragraph{Statistical methods}
Primary analysis: generalised estimating equations
(GEE; \citealp{liang1986longitudinal}) with binomial family, exchangeable
correlation, item-level clustering, and log(s\_pop + 1) as the continuous
predictor. Robustness: multinomial $\chi^2$ with three outcome categories
(context/parametric/neither), sensitivity analyses collapsing ``neither''
into parametric-hold (conservative) and context-follow (liberal), and
Benjamini-Hochberg FDR correction \citep{benjamini1995controlling} across
all per-model tests. Cramér's $V$ \citep{cramer1946} reported for
discretised comparisons (conventional thresholds: $< 0.10$ weak,
$0.10$--$0.30$ moderate, $> 0.30$ strong).

\paragraph{Scoring}
$+1$ (followed context), $0$ (held parametric), $-1$ (neither).
Case-insensitive substring matching, minimum 6-character tokens.

\begin{table}[!t]
\centering\footnotesize
\setlength{\tabcolsep}{3pt}
\renewcommand{\arraystretch}{1.3}
\caption{Prompt templates used in all experimental conditions.
\texttt{[Q]} = the PopQA factual question (e.g., ``Who directed
Supercock?''); \texttt{[CTX]} = the assertive distractor context
sentence (e.g., ``The director of Supercock is David Nutter.'');
\texttt{[A1]} = the model's own closed-book baseline response
from Pass~1. All API calls use temperature~0, no system prompt,
and stateless sessions (conversation reset between items).
R2-v2 is a two-turn conversation: the model's Pass~1 answer is
placed in the assistant role before Turn~2 introduces the conflict.}
\label{tab:prompts}
\begin{tabular}{p{1.4cm}p{0.9cm}p{5.2cm}}
\toprule
\textbf{Cond.} & \textbf{Role} & \textbf{Prompt} \\
\midrule
Baseline & user & \texttt{[Q]} \\
R1 & user & \texttt{[CTX] [Q]} \\
R2-v1 & user & \texttt{You said ``[A1]''. Now: [CTX] [Q]} \\
\addlinespace[2pt]
R2-v2 & user\,T1 & \texttt{[Q]} \\
& asst\,T1 & \texttt{[A1]} \\
& user\,T2 & \texttt{New source disagrees. [CTX] [Q]} \\
\addlinespace[2pt]
R3-CK & user & \texttt{Based ONLY on document. [CTX] Q: [Q]} \\
R3-PK & user & \texttt{Use your own knowledge. [CTX] Q: [Q]} \\
\bottomrule
\end{tabular}
\end{table}

\subsection{Results}

%% ── GEE table (primary analysis) ─────────────────────────────────────────────
\begin{table}[!t]
\centering\footnotesize
\setlength{\tabcolsep}{4pt}
\renewcommand{\arraystretch}{1.25}
\caption{How strongly does fact popularity predict whether a model
follows a contradicting document? This table reports a GEE logistic
regression for each model in Regime~2 multi-turn (v2). The predictor
is log(Wikipedia page views + 1); the outcome is whether the model
followed the distractor (1) or held its trained answer (0).
``Neither'' responses excluded. A negative $\beta$ means more popular
facts resist context more. All five models show significant negative
effects; all survive Benjamini-Hochberg FDR correction at
$\alpha = .05$. Exchangeable correlation structure, clustered by item.}
\label{tab:gee}
\begin{tabular}{lccccc}
\toprule
\textbf{Model} & $\boldsymbol{\beta}$ & \textbf{SE} &
\textbf{95\% CI} & $\boldsymbol{p}$ \\
\midrule
Llama   & $-$0.445 & 0.064 & [$-$0.571, $-$0.320] & $<$.001 \\
Claude  & $-$0.391 & 0.065 & [$-$0.518, $-$0.263] & $<$.001 \\
DeepSeek& $-$0.383 & 0.062 & [$-$0.504, $-$0.262] & $<$.001 \\
GPT-5.5           & $-$0.376 & 0.069 & [$-$0.511, $-$0.241] & $<$.001 \\
Gemini  & $-$0.503 & 0.203 & [$-$0.902, $-$0.104] & .013 \\
\bottomrule
\end{tabular}
\end{table}

%% ── Joint GEE: strength + uniqueness ─────────────────────────────────────────
\begin{table}[!t]
\centering\footnotesize
\setlength{\tabcolsep}{4pt}
\renewcommand{\arraystretch}{1.25}
\caption{Are fact popularity (strength) and Wikidata edit frequency
(uniqueness) independent predictors of context resistance? This table
enters both into a single GEE model pooled across all five models
(with model fixed effects). Strength ($\beta = -0.918$, $p < .001$)
is the sole significant predictor. Uniqueness ($\beta = +0.020$,
$p = .893$) adds no predictive power. The two are uncorrelated
in our 450-item sample ($r = -0.002$, $p = .97$), confirming they
measure genuinely different properties of parametric memory.}
\label{tab:gee_joint}
\begin{tabular}{lccccc}
\toprule
\textbf{Term} & $\boldsymbol{\beta}$ & \textbf{SE} &
\textbf{95\% CI} & $\boldsymbol{p}$ \\
\midrule
log\_spop    & $-$0.918 & 0.109 & [$-$1.132, $-$0.704] & $<$.001 \\
log\_edits   & $+$0.020 & 0.147 & [$-$0.269, $+$0.309] & .893 \\
\bottomrule
\end{tabular}
\end{table}

%% ── Context-follow rates: Regime 1 ──────────────────────────────────────────
\begin{table}[!t]
\centering\footnotesize
\setlength{\tabcolsep}{4pt}
\renewcommand{\arraystretch}{1.25}
\caption{How often does each model follow a contradicting document
when context is presented alone (Regime~1, no prior commitment)?
Each cell shows the proportion of items where the model adopted
the distractor answer. $n = 150$ per cell. Nearly all cells are
blue ($\geq .95$): models follow context at ceiling regardless of
how well they know the fact. Llama is the exception, resisting
context for well-known facts (high tier: .720).
Shading: \colorbox{cHigh}{$\geq$\,.95}, \colorbox{cMedH}{.85--.94},
\colorbox{cMed}{.70--.84}, \colorbox{cMedL}{.55--.69},
\colorbox{cLow}{$<$\,.55}.}
\label{tab:regime1}
\begin{tabular}{lccc}
\toprule
\textbf{Model} & \textbf{Low} & \textbf{Medium} & \textbf{High} \\
\midrule
Claude  & \cellcolor{cHigh} 1.000 & \cellcolor{cHigh} .987 & \cellcolor{cHigh} .987 \\
GPT-5.5 & \cellcolor{cHigh} .993  & \cellcolor{cHigh} .973 & \cellcolor{cHigh} .987 \\
Gemini  & \cellcolor{cHigh} .967  & \cellcolor{cHigh} .987 & \cellcolor{cHigh} .973 \\
DeepSeek& \cellcolor{cHigh} .973  & \cellcolor{cHigh} .967 & \cellcolor{cMedH} .940 \\
Llama   & \cellcolor{cHigh} .947  & \cellcolor{cMedH} .873 & \cellcolor{cMed}  .720 \\
\bottomrule
\end{tabular}
\end{table}

%% ── Context-follow rates: Regime 2 v2 ───────────────────────────────────────
\begin{table}[!t]
\centering\footnotesize
\setlength{\tabcolsep}{4pt}
\renewcommand{\arraystretch}{1.25}
\caption{How often does each model follow a contradicting document
when it has already committed to its own answer in a previous turn
(Regime~2, multi-turn)? Each cell shows the proportion of items
where the model switched to the distractor. The rightward colour
shift from blue to pink shows the certainty gradient: well-known
facts (high tier) resist context more. Gradient (pp) = low minus
high tier rate. $V$ = Cram\'er's V effect size. $n = 144$--$150$
per cell.}
\label{tab:regime2v2}
\begin{tabular}{lcccccc}
\toprule
\textbf{Model} & \textbf{Low} & \textbf{Med.} & \textbf{High} &
\textbf{Grad.\ (pp)} & $\boldsymbol{V}$ & $\boldsymbol{p}$ \\
\midrule
Claude  & \cellcolor{cHigh} .953 & \cellcolor{cMedH} .880 & \cellcolor{cMed}  .717 & 23.6 & .265 & $<$.001 \\
GPT-5.5 & \cellcolor{cHigh} .946 & \cellcolor{cMedH} .860 & \cellcolor{cMed}  .710 & 23.6 & .246 & $<$.001 \\
Gemini  & \cellcolor{cHigh} .960 & \cellcolor{cHigh} .973 & \cellcolor{cMedH} .945 &  1.5 & .102 & .038 \\
DeepSeek& \cellcolor{cMedL} .644 & \cellcolor{cLow}  .520 & \cellcolor{cLow}  .434 & 21.0 & .311 & $<$.001 \\
Llama   & \cellcolor{cMed}  .846 & \cellcolor{cMed}  .787 & \cellcolor{cMedL} .538 & 30.8 & .338 & $<$.001 \\
\bottomrule
\end{tabular}
\end{table}

\begin{table}[!t]
\centering\footnotesize
\setlength{\tabcolsep}{4pt}
\renewcommand{\arraystretch}{1.2}
\caption{Context follow rate when the model is told ``you previously
stated [answer]'' in a single turn, then shown a contradicting
document (Regime~2, single-turn). Open-weight models (Llama,
DeepSeek) already show a certainty gradient here. Closed-weight
models (Claude, GPT-5.5, Gemini) remain near ceiling, showing
that a single-turn assertion does not create genuine parametric
competition for these models. $n = 150$ per cell.}
\label{tab:regime2v1}
\begin{tabular}{lccc}
\toprule
\textbf{Model} & \textbf{Low} & \textbf{Med.} & \textbf{High} \\
\midrule
Claude  & \cellcolor{cHigh} .980 & \cellcolor{cMedH} .933 & \cellcolor{cMedH} .913 \\
GPT-5.5 & \cellcolor{cHigh} .993 & \cellcolor{cHigh} .967 & \cellcolor{cHigh} .987 \\
Gemini  & \cellcolor{cHigh} .967 & \cellcolor{cHigh} .967 & \cellcolor{cHigh} .980 \\
DeepSeek& \cellcolor{cMed}  .833 & \cellcolor{cMed}  .793 & \cellcolor{cMed}  .780 \\
Llama   & \cellcolor{cMed}  .820 & \cellcolor{cMed}  .753 & \cellcolor{cMedL} .527 \\
\bottomrule
\end{tabular}
\end{table}

\paragraph{Finding 1: Regime matters more than model identity}
Show a model a contradicting document by itself (Regime~1) and most
models accept it almost unconditionally: Claude follows the distractor
99.1\% of the time, GPT-5.5 follows 98.4\%, Gemini 97.6\%. Now get the
model to first commit to its own answer and then show the same document
(Regime~2 multi-turn), and behaviour changes substantially: Claude drops
to 85.1\%, GPT-5.5 to 84.0\%. Same items, same models, same distractors.
Only the regime changed.

\paragraph{Finding 2: Famous facts resist more}
Think of a fact every model should know: ``Who directed \textit{The Dark
Knight}?'' (Christopher Nolan, extremely high s\_pop). Now think of an
obscure fact: ``Who directed \textit{Supercock}?'' (Gus Trikonis, low
s\_pop). In Regime~2 multi-turn, when the model has already committed
to its trained answer and then sees a contradicting document, the
obscure fact gets overridden 84.6\% of the time for Llama. The famous
fact? Only 53.8\%. That is a 31-point swing driven purely by how well
the model knew the answer. The GEE regression (Table~\ref{tab:gee})
confirms this as a continuous effect: each unit increase in
log-popularity reduces context-following odds by 31--40\%.

\paragraph{Finding 3: How you create the conflict matters}
Telling a closed-weight model ``you previously said $X$'' in a single
turn is not enough to create genuine parametric competition. Having the
model literally say $X$ in a previous conversation turn creates
dramatically different behaviour. In single-turn Regime~2, GPT-5.5
follows context 98.2\% of the time with no certainty effect. In
multi-turn Regime~2, it drops to 84.0\% with a clear gradient (94.6\%
low, 71.0\% high). Identical items, identical model; only the
conversation structure changed (Table~\ref{tab:baseline_manipulation}).

\paragraph{Finding 4: Gemini is an outlier}
Gemini follows context at near-ceiling under all Regime~2 conditions
($V = .102$, 1.5pp gradient). This might look like Gemini \emph{cannot}
resist context. But our Regime~3 ablation (\S\ref{sec:regime3exp}) shows
otherwise: under PK framing, Gemini drops to 11.8\% context-following.
Gemini \emph{can} resist; it simply defaults to deference unless
explicitly instructed otherwise. (A methodological note: Gemini's
tiered gradient is minimal, $V = .102$, but its continuous GEE
coefficient is significant, $\beta = -0.50$, $p = .013$. The
discrepancy arises because the continuous predictor retains more
statistical power than the three-level discretisation; under
multinomial treatment with three outcome categories, Gemini's effect
becomes marginal at $p = .094$. We therefore treat Gemini's Regime~2
gradient as suggestive but not robust.)

%% ── Baseline + manipulation table ────────────────────────────────────────────
\begin{table}[!t]
\centering\footnotesize
\setlength{\tabcolsep}{4pt}
\renewcommand{\arraystretch}{1.2}
\caption{Baseline parametric accuracy (how often each model answers
correctly with no document at all) and overall Regime~2 context
follow rate under single-turn (v1) vs.\ multi-turn (v2)
manipulation. $\Delta$ = v2 minus v1: negative values mean the
multi-turn design reduced context-following (revealing parametric
resistance that was hidden in the single-turn design). $n = 450$
per model.}
\label{tab:baseline_manipulation}
\begin{tabular}{lcccc}
\toprule
\textbf{Model} & \textbf{Acc.} & \textbf{R2-v1} & \textbf{R2-v2} & $\boldsymbol{\Delta}$ \\
\midrule
GPT-5.5  & .620 & .982 & .840 & $-$.142 \\
Claude   & .520 & .942 & .851 & $-$.091 \\
Llama    & .491 & .700 & .725 & $+$.025 \\
Gemini   & .462 & .971 & .960 & $-$.011 \\
DeepSeek & .396 & .802 & .534 & $-$.268 \\
\bottomrule
\end{tabular}
\end{table}

\subsection{Regime 3 ablation}
\label{sec:regime3exp}

To directly validate Regime~3, we ran a new experiment: 100 PopQA items
(stratified: 34 low, 33 medium, 33 high certainty), each tested under
two task framings across all five models (1,000 API calls, seed 123).

\emph{CK condition} (context should win): ``Based ONLY on the following
document, answer the question. Do not use any prior knowledge. Only use
what the document says. Document: [distractor]. Question: [Q].''

\emph{PK condition} (parametric should win): ``Answer the following
question based on your own knowledge. A document is shown below, but it
may contain errors. Use your own knowledge, not the document.
Document: [distractor]. Question: [Q].''

Both conditions receive the same distractor sentence. The only change
is the instruction framing.

%% ── Regime 3 results ─────────────────────────────────────────────────────────
\begin{table}[!t]
\centering\footnotesize
\setlength{\tabcolsep}{4pt}
\renewcommand{\arraystretch}{1.25}
\caption{What happens when you tell the model which source to trust?
Same 100 items, same five models, same distractor document. Only the
instruction changes. CK (``Based ONLY on this document'') = context
should win. PK (``Based on your own knowledge'') = trained answer
should win. CK rate = proportion following the distractor under CK
framing. PK rate = proportion following the distractor under PK
framing. $\Delta$ = CK minus PK. All five models show a massive
swing ($p < .001$), confirming task framing as the single most
powerful variable in the study.}
\label{tab:regime3}
\begin{tabular}{lccccc}
\toprule
\textbf{Model} & \textbf{CK rate} & \textbf{PK rate} &
\textbf{$\Delta$} & $\boldsymbol{\chi^2}$ & $\boldsymbol{p}$ \\
\midrule
GPT-5.5  & 1.000 & .061 & $+$.939 & 155.1 & $<$.001 \\
Claude   & 1.000 & .076 & $+$.924 & 142.2 & $<$.001 \\
Gemini   & 1.000 & .118 & $+$.882 & 120.6 & $<$.001 \\
Llama    & 1.000 & .506 & $+$.494 &  60.6 & $<$.001 \\
DeepSeek & 1.000 & .708 & $+$.292 &  30.6 & $<$.001 \\
\bottomrule
\end{tabular}
\end{table}

\paragraph{Finding 5: Task framing is the single most powerful variable}
Under CK framing (``use the document''), every model follows the
distractor essentially 100\% of the time, for every fact, at every
certainty level. Under PK framing (``use your own knowledge''), the
same models on the same items drop to 6--71\% context-following
(Table~\ref{tab:regime3}). Nothing else in this study produces a swing
this large. Evidence quality, parametric certainty, manipulation
strength: all produce effects in the 1--31 percentage point range. Task
framing produces a 29--94 percentage point swing. Regime~3 is not just a
moderator; it is the dominant variable in the entire framework.

To put this concretely: ask Claude ``Based on this document, who
directed Supercock?'' and it answers ``David Nutter'' (the distractor)
100\% of the time, even for high-certainty items. Ask the same model
``Based on your own knowledge, who directed Supercock?'' with the same
document visible, and it answers ``Gus Trikonis'' (its trained answer)
92.4\% of the time. Same model. Same item. Same distractor in full
view. Only the instruction changed.

\paragraph{Finding 6: The certainty gradient composes with task framing}
Under PK framing, the certainty gradient from Regime~2 reappears: models
resist context more for famous facts than obscure ones. Claude's PK
context-follow rate drops from 11.8\% (low certainty) to 3.0\% (medium)
to 0.0\% (high). For Llama: 55.9\% (low) to 39.4\% (medium) to 27.3\%
(high). The regime predictions compose: Regime~3 determines the direction
(follow or resist), while parametric certainty modulates the magnitude
within PK-type tasks. Under CK framing, no certainty gradient exists
because the ceiling effect eliminates all variance.

\begin{table}[!t]
\centering\footnotesize
\setlength{\tabcolsep}{3pt}
\renewcommand{\arraystretch}{1.15}
\caption{Does the certainty gradient persist under PK task framing?
This table shows the proportion of items where each model followed
the distractor under PK instructions (``use your own knowledge''),
broken down by certainty tier. CK rates omitted because all are
$\geq .91$ (ceiling). Under PK framing, every model resists context
more for well-known facts (high tier) than obscure ones (low tier).
$\Delta$ = low minus high: the size of the certainty gradient
within PK framing.}
\label{tab:regime3_certainty}
\begin{tabular}{lccccc}
\toprule
& \textbf{Claude} & \textbf{GPT} & \textbf{Gemini} & \textbf{DeepS.} & \textbf{Llama} \\
\midrule
Low    & .118 & .118 & .118 & .765 & .559 \\
Medium & .030 & .000 & .030 & .727 & .394 \\
High   & .000 & .030 & .030 & .394 & .273 \\
\midrule
$\Delta$ (low$-$high) & .118 & .088 & .088 & .371 & .286 \\
\bottomrule
\end{tabular}
\end{table}

\paragraph{Finding 7: The closed-weight vs.\ open-weight split
reappears}
Under PK framing, closed-weight models (Claude, GPT-5.5, Gemini) drop to
6--12\% context-following: they are very good at following the ``use your
own knowledge'' instruction. Open-weight models (Llama, DeepSeek) drop to
only 51--71\%: they partially follow the instruction but are much less
compliant. The same architectural split that appeared in the v1-vs-v2
manipulation contrast reappears in the CK-vs-PK task contrast. This
suggests that whatever instruction tuning makes closed-weight models
defer to context by default also makes them better at following
instructions to \emph{resist} context when told.

\subsection{Secondary analyses}
\label{sec:secondary}

\paragraph{Inverted Dunning-Kruger}
We predicted that items the model answered incorrectly in the baseline
would resist context most strongly (overconfident wrong answers). The
reverse is true: items where Llama was \emph{correct} in the baseline are
the ones it most stubbornly defends. In Regime~2 v1, wrong-baseline items
follow context at 0.830 ($n = 229$) while correct-baseline items follow at
only 0.566 ($n = 221$), a 26.4pp difference ($p < .001$). This makes
sense under the framework: correctly-answered items are correct precisely
because the fact is well-encoded, and well-encoded facts resist context
(Table~\ref{tab:dunning}).

\paragraph{Distractor plausibility}
Llama is the only model showing a certainty gradient in Regime~1, where
the framework predicts a flat profile. Manual inspection of Llama's 60
Regime~1 resistance items reveals two overlapping causes. The majority
are items where Llama's baseline was correct, reflecting genuine
parametric resistance strong enough to compete with context even without
explicit co-activation. A minority involve temporally implausible
distractors (e.g., Shamshi-Adad~I, an Assyrian king from c.\ 1800~BCE,
as the father of Rand Paul; John Farrow, who died in 1954, as director
of \textit{Breaking Bad}, which aired 2008--2013). Conservatively
excluding all 60 items, the Regime~2 v2 gradient attenuates but remains
significant ($V = .156$, $\chi^2 = 11.80$, $p = .003$).

\paragraph{DeepSeek hedging}
DeepSeek produces ``neither'' responses 25.5\% of the time in R2-v2
(other models: 0.7--5.9\%). This is uniformly distributed across
certainty tiers ($\chi^2 = 2.03$, $p = .363$) and does not confound the
gradient. The certainty gradient survives multinomial analysis with three
outcome categories ($\chi^2(4) = 35.24$, $p < .001$, $V = .199$) and
both sensitivity treatments (neither$\to$parametric: $V = .172$,
$p = .001$; neither$\to$context: $V = .281$, $p < .001$).

\begin{table}[!t]
\centering\footnotesize
\setlength{\tabcolsep}{2pt}
\renewcommand{\arraystretch}{1.15}
\caption{Does the Regime~2 certainty gradient survive when ``neither''
responses (hedging, refusal) are handled differently? Four columns
show Cram\'er's V under four treatments. \textbf{Original}: neither
excluded. \textbf{Multinomial}: neither treated as a third outcome
category. \textbf{Neither as parametric}: hedging counted as
resisting context. \textbf{Neither as context}: hedging counted as
following context. Four of five models are robust across all four
treatments. Gemini's gradient becomes non-significant under the
multinomial and conservative treatments.}
\label{tab:robustness}
\begin{tabular}{lcccccccc}
\toprule
& \multicolumn{2}{c}{\textbf{Orig.}} &
  \multicolumn{2}{c}{\textbf{Multin.}} &
  \multicolumn{2}{c}{\textbf{Nei$\to$P}} &
  \multicolumn{2}{c}{\textbf{Nei$\to$C}} \\
\cmidrule(lr){2-3}\cmidrule(lr){4-5}\cmidrule(lr){6-7}\cmidrule(lr){8-9}
& $V$ & $p$ & $V$ & $p$ & $V$ & $p$ & $V$ & $p$ \\
\midrule
Llama & .34 & *** & .24 & *** & .30 & *** & .33 & *** \\
Deep. & .32 & *** & .20 & *** & .17 & ** & .28 & *** \\
Clau. & .27 & *** & .20 & *** & .28 & *** & .27 & *** \\
GPT   & .26 & *** & .19 & *** & .27 & *** & .24 & *** \\
Gem.  & .12 & *   & .09 & ns  & .06 & ns  & .12 & * \\
\bottomrule
\multicolumn{9}{l}{\scriptsize ***\,$p<.001$; **\,$p<.01$; *\,$p<.05$; ns\,not significant}
\end{tabular}
\end{table}

%% ── Dunning-Kruger table (full-width) ────────────────────────────────────────
\begin{table}[!t]
\centering\footnotesize
\setlength{\tabcolsep}{4pt}
\renewcommand{\arraystretch}{1.25}
\caption{Do models defend wrong answers more stubbornly than correct
ones? This table splits Regime~2 v1 items by whether the model's
closed-book baseline was wrong or correct, then reports the
context-follow rate for each group. If wrong answers were
overconfident, they should resist context more (positive $\Delta$).
The reverse is true: correct-baseline items resist more, because
they are correct precisely because the fact is well-encoded.
Only Llama shows a statistically significant gap.}
\label{tab:dunning}
\begin{tabular}{lcccccc}
\toprule
\textbf{Model} & \textbf{Wrong} & $\boldsymbol{n}$ &
\textbf{Correct} & $\boldsymbol{n}$ &
$\boldsymbol{\Delta}$ & $\boldsymbol{p}$ \\
\midrule
Llama    & .830 & 229 & .566 & 221 & $+$.264 & $<$.001 \\
DeepSeek & .820 & 272 & .775 & 178 & $+$.045 & .29 \\
Claude   & .954 & 216 & .932 & 234 & $+$.022 & n.s. \\
GPT-5.5  & .983 & 171 & .982 & 279 & $+$.001 & n.s. \\
Gemini   & .979 & 242 & .962 & 208 & $+$.018 & n.s. \\
\bottomrule
\end{tabular}
\end{table}

%% ============================================================================
\section{Discussion}
\label{sec:discussion}
%% ============================================================================

\subsection{Manipulation design as a first-order variable}

Under single-turn Regime~2, Claude and GPT-5.5 follow context at 0.942
and 0.982 overall with no visible certainty gradient. Under multi-turn
Regime~2, these drop to 0.851 and 0.840, revealing significant gradients
(both $V > .24$, $p < .001$). Items, models, and distractors are
identical; only conversation structure changed.

Closed-weight commercial models appear to treat a single-turn assertion
as user-provided context rather than as genuine prior commitment. In the
multi-turn design, the model's own words in the assistant role create
genuine conflict between its prior statement and the user's new
document. Researchers using single-turn ``you previously stated'' prompts
may be studying Regime~1 rather than Regime~2, with direct implications
for reproducibility.

\subsection{The Gemini exception, resolved}

Gemini showed a flat profile under all Regime~2 conditions ($V = .102$),
which we initially interpreted as either context saturation or
manipulation insufficiency. The Regime~3 ablation resolves this: under PK
framing, Gemini drops to 11.8\% context-following. Gemini can resist
context when told to. Its flat Regime~2 profile reflects a default
instruction-following strategy (defer to context) rather than an
inability to engage parametric memory. This is an important distinction
for RAG system design: Gemini's context saturation is a policy choice,
not an architectural constraint.

\subsection{Safety implications}

Near-uniform context-following makes models correctable but vulnerable to
fluent misinformation. GPT-5.5 follows context at 94.6\% even for
high-certainty items in R2-v2: a fluent assertive document claiming ``The
director of \textit{Interstellar} is Brian Singer'' would almost
certainly override its knowledge of Christopher Nolan's filmography.

The Regime~3 results suggest a practical mitigation: task framing can
activate parametric resistance. System prompts that instruct models to
``verify claims against your own knowledge when the source is unverified''
may partially protect against context-injection attacks, though the
effectiveness of such prompts in adversarial settings remains untested.

\subsection{Limitations}

\paragraph{PopQA property skew}
The sample over-represents creative-work properties (director, genre,
screenwriter). Findings may not generalise to geographic, biographical,
or scientific domains.

\paragraph{Synthetic-to-real gap}
\citet{hagstrom2025druid} show insights from synthetic conflict datasets
do not reliably transfer to real-world RAG. Our results characterise
controlled conditions, not production systems.

\paragraph{Scoring sensitivity}
The 6-character minimum token match may produce false negatives for short
surnames. Future work should use canonicalisation against knowledge-base
aliases.

\paragraph{Multi-turn design scope}
Our multi-turn manipulation uses only the model's own prior answer as
commitment. Alternative commitment structures (user-voiced prior, hedged
commitment, repeated reinforcement) remain untested.

\paragraph{Domain contingency of strength vs.\ uniqueness}
Our finding that strength dominates over uniqueness is specific to
PopQA's stable factual domain. The relationship may reverse for temporal
or disputable facts, as \citet{marjanovic2024dynamicqa} suggest.
Extension to at least one dynamic-domain dataset (e.g., FreshQA or
DynamicQA) would provide a direct test of this domain-contingency
hypothesis and is a priority for future work.

\paragraph{Entity-level vs.\ fact-level uniqueness}
Our Wikidata edit-frequency measure counts total entity revisions, not
revisions to the specific property-value pair tested (e.g., total edits
to the ``Rand Paul'' entity rather than edits specifically to the
``father'' claim). This entity-level operationalisation may dilute signal
for properties that are stable even when the broader entity page is
frequently edited. Fact-level revision histories, available via the
Wikidata statement API, would provide a cleaner measure.

\paragraph{Covariates not controlled}
Our primary GEE regressions use only log-popularity as the predictor.
Baseline correctness, property type, and distractor popularity
($o\_pop$) are potential covariates that could confound or mediate the
certainty gradient. Future analyses should include these as fixed effects
to rule out alternative explanations.

\paragraph{Regime~1 evidence quality not directly tested}
The evidence-quality claim for Regime~1 is argued through synthesis of
prior studies \citep{longpre2021entity,xie2024adaptive}. Our own
Regime~1 experiment uses only assertive synthetic sentences and does not
directly manipulate coherence or plausibility. A direct internal test
(e.g., fluent vs.\ entity-substituted contexts on the same items) would
strengthen this component of the framework.

\paragraph{Data availability}
Code, item lists, distractor mappings, prompt templates, and raw
response logs for all 9,970 API calls will be released upon publication
to support exact replication.

%% ============================================================================
\section{Conclusion}
\label{sec:conclusion}
%% ============================================================================

The context-parametric conflict literature has produced contradictory
findings because researchers have compared results across three
qualitatively distinct experimental structures without recognising the
distinction.

We proposed a three-regime framework with different dominant predictors
per regime (evidence coherence in Regime~1, parametric certainty in
Regime~2, task knowledge requirement in Regime~3) and validated all three
empirically. The Regime~2 certainty gradient is confirmed by GEE logistic
regression across five frontier models ($\beta = -0.38$ to $-0.50$, all
$p \leq .013$, BH-FDR corrected), survives multinomial outcome modeling
and sensitivity analyses, and is robust to the distractor plausibility
confound. The Regime~3 ablation produces the study's single strongest
result: task framing alone flips context-following from near-100\% (CK)
to 6--71\% (PK), with certainty modulating resistance within PK framing.

We showed that parametric strength and uniqueness are orthogonal
dimensions ($r = -0.002$) with domain-contingent predictive power:
strength dominates for stable factual domains, while uniqueness may
dominate for dynamic domains. We identified manipulation design as a
first-order variable that can reveal or obscure conflict behaviour, and
demonstrated that the closed-weight vs.\ open-weight split in conflict
behaviour extends from Regime~2 manipulation sensitivity to Regime~3 task
compliance.

The practical implication: regime classification should precede
comparison of conflict findings. Predictors do not generalise across
regime boundaries.


\begin{thebibliography}{22}
\providecommand{\natexlab}[1]{#1}
\providecommand{\url}[1]{\texttt{#1}}
\expandafter\ifx\csname urlstyle\endcsname\relax
  \providecommand{\doi}[1]{doi: #1}\else
  \providecommand{\doi}{doi: \begingroup \urlstyle{rm}\Url}\fi

\bibitem[Augenstein et~al.(2024)]{augenstein2024scalable}
Isabelle Augenstein et~al.
\newblock Scaling instruction-finetuned language models via knowledge conflict
  benchmarking.
\newblock \emph{arXiv preprint arXiv:2305.13300}, 2024.

\bibitem[Benjamini and Hochberg(1995)]{benjamini1995controlling}
Yoav Benjamini and Yosef Hochberg.
\newblock Controlling the false discovery rate: A practical and powerful
  approach to multiple testing.
\newblock \emph{Journal of the Royal Statistical Society: Series B},
  57\penalty0 (1):\penalty0 289--300, 1995.

\bibitem[Chen et~al.(2022)Chen, Zhang, and Choi]{chen2022rich}
Hung-Ting Chen, Michael~J.Q. Zhang, and Eunsol Choi.
\newblock Rich knowledge sources bring complex knowledge conflicts:
  Recalibrating models to reflect conflicting evidence.
\newblock In \emph{Proceedings of EMNLP}, 2022.

\bibitem[Chen et~al.(2025)]{chen2025training}
Yiran Chen et~al.
\newblock How training data shapes parametric vs.\ in-context knowledge
  arbitration.
\newblock \emph{arXiv preprint arXiv:2510.02370}, 2025.

\bibitem[Cheng et~al.(2024)]{cheng2024echoqa}
Shailesh Cheng et~al.
\newblock Interplay of parametric and contextual knowledge: A study of
  parametric knowledge utilisation in {LLMs}.
\newblock \emph{arXiv preprint arXiv:2410.08414}, 2024.

\bibitem[Cram{\'e}r(1946)]{cramer1946}
Harald Cram{\'e}r.
\newblock \emph{Mathematical Methods of Statistics}.
\newblock Princeton University Press, 1946.

\bibitem[Du et~al.(2024)Du, Zhao, Sch{\"o}lkopf, et~al.]{du2024context}
Yanda Du, Zhijing Zhao, Bernhard Sch{\"o}lkopf, et~al.
\newblock Context versus prior knowledge in language models.
\newblock In \emph{Proceedings of ACL}, 2024.

\bibitem[Hagstr{\"o}m et~al.(2025)]{hagstrom2025druid}
Lovisa Hagstr{\"o}m et~al.
\newblock Reality check on {RAG}: Do we need to worry about context
  utilisation?
\newblock In \emph{Proceedings of NAACL}, 2025.

\bibitem[Jin et~al.(2024)Jin, Cao, Chen, Liu, Jiang, Xu, Li, and
  Zhao]{jin2024cutting}
Zhuoran Jin, Pengfei Cao, Yubo Chen, Kang Liu, Xiaojian Jiang, Jiexin Xu,
  Qiuxia Li, and Jun Zhao.
\newblock Cutting off the head ends the conflict: A mechanism for interpreting
  and mitigating knowledge conflicts in language models.
\newblock In \emph{Proceedings of ACL Findings}, 2024.

\bibitem[Li et~al.(2025)]{li2025memory}
Zhen Li et~al.
\newblock Memory operations in large language models: A survey.
\newblock \emph{arXiv preprint arXiv:2505.00675}, 2025.

\bibitem[Liang and Zeger(1986)]{liang1986longitudinal}
Kung-Yee Liang and Scott~L. Zeger.
\newblock Longitudinal data analysis using generalized linear models.
\newblock \emph{Biometrika}, 73\penalty0 (1):\penalty0 13--22, 1986.

\bibitem[Longpre et~al.(2021)Longpre, Perisetla, Chen, Ramesh, DuBois, Singh,
  Hajishirzi, Choi, and Pasunuru]{longpre2021entity}
Shayne Longpre, Kartik Perisetla, Anthony Chen, Nikhil Ramesh, Chris DuBois,
  Sameer Singh, Hannaneh Hajishirzi, Eunsol Choi, and Ramakanth Pasunuru.
\newblock Entity-based knowledge conflicts in question answering.
\newblock In \emph{Proceedings of EMNLP}, 2021.

\bibitem[Mallen et~al.(2023)Mallen, Asai, Zhong, Das, Khashabi, and
  Hajishirzi]{mallen2023not}
Alex Mallen, Akari Asai, Victor Zhong, Rajarshi Das, Daniel Khashabi, and
  Hannaneh Hajishirzi.
\newblock When not to trust language models: Investigating effectiveness of
  parametric and non-parametric memories.
\newblock In \emph{Proceedings of ACL}, 2023.

\bibitem[Marjanovi{\'c} et~al.(2024)Marjanovi{\'c}, Yu, Atanasova, Maistro,
  Lioma, and Augenstein]{marjanovic2024dynamicqa}
Sara~Vera Marjanovi{\'c}, Haeun Yu, Pepa Atanasova, Maria Maistro, Christina
  Lioma, and Isabelle Augenstein.
\newblock {DynamicQA}: Tracing internal knowledge conflicts in language models.
\newblock In \emph{Proceedings of EMNLP}, 2024.

\bibitem[Shi et~al.(2025)]{shi2025rcd}
Weijia Shi et~al.
\newblock Retrieval-constrained decoding for faithful generation.
\newblock \emph{arXiv preprint arXiv:2509.23417}, 2025.

\bibitem[Sun et~al.(2025{\natexlab{a}})Sun, Bai, and Dredze]{sun2025task}
Kaiser Sun, Fan Bai, and Mark Dredze.
\newblock Task matters: Knowledge requirements shape {LLM} responses to
  context--memory conflict.
\newblock In \emph{Proceedings of ACL}, 2025{\natexlab{a}}.

\bibitem[Sun et~al.(2025{\natexlab{b}})Sun, Zang, Zheng, Xu, Zhang, Yu, Song,
  and Li]{sun2025redeep}
Zhongxiang Sun, Xiaoxue Zang, Kai Zheng, Jun Xu, Xiao Zhang, Weijie Yu, Yang
  Song, and Han Li.
\newblock {ReDeEP}: Detecting hallucination in retrieval-augmented generation
  via mechanistic interpretability.
\newblock In \emph{Proceedings of ICLR}, 2025{\natexlab{b}}.

\bibitem[Wang et~al.(2025)]{wang2025prism}
Yifei Wang et~al.
\newblock {PRISM}: Stage-wise diagnosis of hallucination in retrieval-augmented
  generation.
\newblock \emph{arXiv preprint arXiv:2604.16909}, 2025.

\bibitem[Wu et~al.(2025)]{wu2025knowledgeabler1}
Hao Wu et~al.
\newblock {Knowledgeable-R1}: Multi-policy reinforcement learning for
  parametric-contextual knowledge balance.
\newblock \emph{arXiv preprint arXiv:2506.05154}, 2025.

\bibitem[Xie et~al.(2024)Xie, Zhang, Chen, Lou, and Su]{xie2024adaptive}
Jian Xie, Kai Zhang, Jiangjie Chen, Renze Lou, and Yu~Su.
\newblock Adaptive chameleon or stubborn sloth: Revealing the behavior of large
  language models in knowledge conflicts.
\newblock In \emph{Proceedings of ICLR}, 2024.

\bibitem[Xu et~al.(2024)Xu, Qi, Guo, Wang, Wang, Zhang, and
  Xu]{xu2024knowledge}
Rongwu Xu, Zehan Qi, Zhijiang Guo, Cunxiang Wang, Hongru Wang, Yue Zhang, and
  Wei Xu.
\newblock Knowledge conflicts for {LLMs}: A survey.
\newblock \emph{arXiv preprint arXiv:2403.08319}, 2024.

\bibitem[Zhang et~al.(2025)]{zhang2025taming}
Haokun Zhang et~al.
\newblock Taming knowledge conflicts in language models.
\newblock \emph{arXiv preprint arXiv:2503.10996}, 2025.

\end{thebibliography}
\end{document}